\documentclass[10pt,twocolumn,letterpaper]{article}

\usepackage{iccv}
\usepackage{times}
\usepackage{epsfig}
\usepackage{graphicx}
\usepackage{amsmath}
\usepackage{amssymb}

\usepackage{blindtext}
\usepackage{caption}
\usepackage{amsfonts}
\usepackage[inline]{enumitem}

\usepackage{algorithmic}
\usepackage{textcomp}

\usepackage{booktabs}
\usepackage{multirow}
\usepackage{float}
\usepackage{adjustbox}
\usepackage{etoolbox}
\usepackage{dsfont}

\usepackage{float}

\usepackage{lipsum}
\usepackage{stfloats}
\usepackage{multicol}
\usepackage{amsmath,bm}

\usepackage{array}

\usepackage{tabulary}
\usepackage[table]{xcolor}

\usepackage{paralist}
\usepackage[ruled,linesnumbered]{algorithm2e}

\newcommand{\gray}[1]{\textcolor{gray}{#1}}
\newcommand{\green}[1]{\textcolor[RGB]{96,177,87}{#1}}
\newcommand{\fn}[1]{\footnotesize{#1}}
\newcommand{\gbf}[1]{\green{\bf{\fn{(#1)}}}}
\newcommand{\rbf}[1]{\gray{\bf{\fn{(#1)}}}}

\definecolor{gray}{rgb}{0.3,0.3,0.3}
\definecolor{blue}{rgb}{0,0.5,1}
\definecolor{red}{rgb}{1,0,0.8}
\definecolor{green}{rgb}{0.2,1,0.2}

\definecolor{rblue}{rgb}{0,0,1}
\definecolor{rred}{rgb}{1,0,0}

\let\oldtwocolumn\twocolumn
\renewcommand\twocolumn[1][]{%
    \oldtwocolumn[{#1}{
    \begin{center}
    \vskip-5ex
        \centering
        \includegraphics[width=0.99\textwidth]{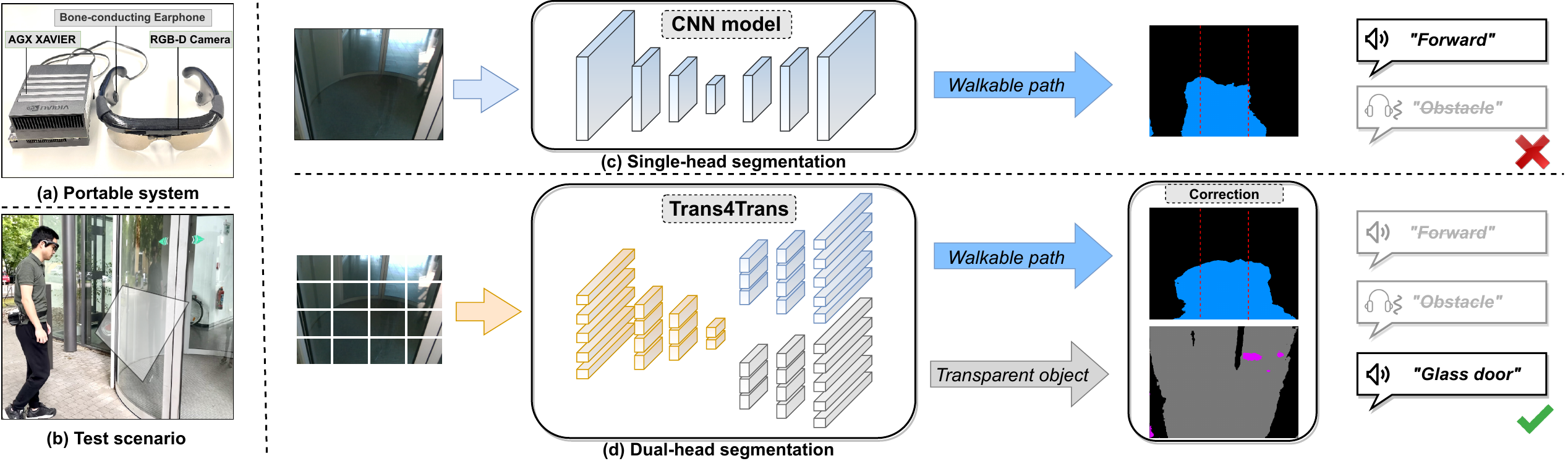}
        \vskip-2ex
        \captionof{figure}{\small (a) the assistive system 
        equipped with smart vision glasses and a portable GPU
        is tested (b) in front of a glass door. The input image is segmented as \textbf{\textcolor{blue}{\emph{Walkable Path}}} by (c) a single head CNN model, and is corrected as \textbf{\textcolor{gray}{\emph{Glass Door}}} by (d) our Transformer for Transparency~(Trans4Trans) model, which is safety-critical for navigation. The user interface consists of vibration and voice feedback.}
        \label{fig1:sys}
    \end{center}
    }]
}

\usepackage[pagebackref=true,breaklinks=true,letterpaper=true,colorlinks,bookmarks=false]{hyperref}

\iccvfinalcopy 

\usepackage[pagebackref=true,breaklinks=true,colorlinks,bookmarks=false]{hyperref}
\hypersetup{colorlinks, citecolor=rblue}
\hypersetup{colorlinks, linkcolor=rred}

\def\iccvPaperID{13} 

\ificcvfinal\fi

\begin{document}
\title{Trans4Trans: Efficient Transformer for Transparent Object Segmentation to Help Visually Impaired People Navigate in the Real World}

\author{Jiaming Zhang, Kailun Yang\thanks{Corresponding author (e-mail: {\tt kailun.yang@kit.edu}).}, Angela Constantinescu, Kunyu Peng, Karin M\"uller and Rainer Stiefelhagen\\
Karlsruhe Institute of Technology
}

\maketitle
\ificcvfinal\thispagestyle{empty}\fi

\begin{abstract}
Common fully glazed facades and transparent objects present architectural barriers and impede the mobility of people with low vision or blindness, for instance, a path detected behind a glass door is inaccessible unless it is correctly perceived and reacted. However, segmenting these safety-critical objects is rarely covered by conventional assistive technologies. To tackle this issue, we construct a wearable system with a novel dual-head Transformer for Transparency \emph{(Trans4Trans)} model, which is capable of segmenting general and transparent objects and performing real-time wayfinding to assist people walking alone more safely. Especially, both decoders created by our proposed Transformer Parsing Module \emph{(TPM)} enable effective joint learning from different datasets. Besides, the efficient Trans4Trans model composed of symmetric transformer-based encoder and decoder, requires little computational expenses and is readily deployed on portable GPUs. Our Trans4Trans model outperforms state-of-the-art methods on the test sets of Stanford2D3D and Trans10K-v2 datasets and obtains mIoU of 45.13\% and 75.14\%, respectively. Through various pre-tests and a user study conducted in indoor and outdoor scenarios, the usability and reliability of our assistive system have been extensively verified.

\end{abstract}

\section{Introduction}
Knowledge of glass architecture~\cite{butera2005glass} and glass doors~\cite{maringer2019suitability,mei2020don} are particular important for visually impaired people, because transparent objects often present architectural barriers which hinder the mobility of people with low vision or blindness. For example, a path behind a glass door is not a free way to navigate (see Fig.~\ref{fig1:sys}) unless it is correctly recognized and reacted. However, most common vision-based navigation assistance systems~\cite{aladren2014navigation,wang2017enabling,yang2017ir} cannot handle transparent obstacles well, as 3D vision-based methods hardly recover the depth information of texture-less transparent surfaces~\cite{aladren2014navigation,yang2017ir}, whereas conventional image segmentation-based methods do not cover the categories of challenging transparent objects~\cite{lin2019deep,yang2018unifying}. In addition, guide dogs often get confused leading people with blindness to full-pane windows, and differentiation between doors, and large glass windows is difficult for people with residual sight~\cite{saha2019wayfinding}.
A system that supports the recognition of landmarks such as doors is particularly appreciated by people with visual impairments, as finding a door before entering a building is difficult due to the inaccuracy of GPS~\cite{berenguel2020floor,saha2019wayfinding}.

To address these issues, we propose a wearable system capable of real-time wayfinding and object segmentation to assist visually impaired individuals travel more safely. We present \emph{Trans4Trans}, precisely \emph{Transformer for Transparency}, an efficient semantic segmentation architecture with dual heads, as shown in Fig.~\ref{fig1:sys}(d). As transparent objects are often texture-less or share similar content as the surroundings, it is essential to associate long-range visual concepts to robustly infer transparent regions. For this reason, Trans4Trans is established with both transformer-based encoder and decoder to fully exploit the long-range context modeling capacity of self-attention layers in transformers~\cite{vaswani2017attention}. In particular, Trans4Trans features a novel \emph{Transformer Paring Module (TPM)} to fuse multi-scale feature maps generated from embeddings of dense partitions, and the symmetric transformer-based decoder can consistently parse the feature maps from transformer-based encoder.
Together with semantically predicting general things and stuff classes like walkable areas, the dual-head design allows to segment transparent objects accurately and completely, which are safety-critical for navigation.

Trans4Trans is integrated in our wearable system which comprises a pair of smart vision glasses and a mobile GPU processor, which delivers a holistic scene understanding swiftly and accurately thanks to the high efficiency of our model. 
With the complete semantic information, the user interface consists of a customized set of acoustic feedback via sonification of detected objects, walkable directions and warnings of the obstacles, which yields intuitive suggestions and no prior knowledge is needed.
A comprehensive set of experiments has been conducted on multiple semantic segmentation datasets~\cite{stanford2d3d,xie2021segmenting}. In particular, the proposed model outperforms state-of-the-art methods on the test sets of Stanford2D3D~\cite{stanford2d3d} and Trans10K-v2~\cite{xie2021segmenting} datasets. Finally, a user study with visually impaired people and a variety of field tests demonstrate the usability and reliability of our assistive system for navigational perception in the wild. To the best of our knowledge, we are the first to use vision transformers for assisting people with visual impairment.

In summary, we deliver the following contributions:
\begin{compactitem}
    \item We present a wearable assistive system with a pair of smart vision glasses and a mobile GPU ported with vision transformers for visually impaired people.
    \item We propose an efficient semantic segmentation architecture Transformer for Transparency (\emph{Trans4Trans}) with transformer-based encoder and decoder, a dual-head design to unify general object and challenging transparent object segmentation, and a Transformer Parsing Module to fuse multi-scale representations.
    \item Trans4Trans, maintaining a high efficiency, surpasses state-of-the-art CNN- and transformer-based methods on Stanford2D3D and Tran10K-v2 datasets.
    \item According to our designed system algorithm, we produce a customized set of acoustic feedback and conduct a user study and various field tests, demonstrating the usability and reliability of Trans4Trans. 
\end{compactitem}

\section{Related Work}

\noindent\textbf{Semantic segmentation for visual assistance.}
Whereas traditional assistance systems rely on multiple monocular detectors and depth sensors~\cite{aladren2014navigation,cheng2018real,cheng2017crosswalk,lin2018krnet,wang2017enabling}, semantic segmentation allows to solve many navigational perception problems at once and thereby has been quickly employed in visual assistance.
Yang~\etal~\cite{yang2018unifying} put forward seizing semantic segmentation to unify detection tasks and assist terrain awareness, whereas Mao~\etal~\cite{mao2021panoptic} argued for panoptic segmentation towards a holistic sensing. In~\cite{long2019unifying,yohannes2019content}, instance-specific segmentation methods like Mask R-CNN~\cite{he2017mask} were directly applied for content-aware surrounding understanding.
Semantic segmentation has also been used to address intersection perception like detection of crosswalks, sidewalks, and blind roads~\cite{cao2020rapid,hsieh2020outdoor}. Moreover, it has received increasing interests and various systems appear in the field~\cite{constantinescu2020bring,duh2020v,hsieh2020development,lin2019deep,miksik2015semantic}. Yet, both traditional sensor-based and segmentation-driven approaches cannot handle challenging transparent obstacles well.

\noindent\textbf{Transparent object sensing.}
Classical visual assistance systems~\cite{bai2017smart,huang2018glass} resort to multi-sensor fusion, \eg, fusing RGB-D cameras and ultrasonic sensors, to overcome the difficulties in dealing with transparent obstacles like glass objects, French windows, French doors, etc. Chen~\etal~\cite{chen2018improving} design a multimodal stereo matching algorithm to improve the depth measurements of transparent objects with dual depth sensors. Polarization cues~\cite{deep_polarization_cues} and reflection priors~\cite{lin2021rich} are also frequently explored for transparency perception. For example, Xiang~\etal~\cite{xiang2021polarization} propose a polarization-driven semantic segmentation architecture by adaptively bridging RGB and polarization dimensions,
which significantly lifts the performance of classes with polarization properties like \emph{glass}.

Recently, Xie~\etal~\cite{xie2020segmenting,xie2021segmenting} built the Trans10K dataset and show that while the pure RGB-based transparent object segmentation is a largely unsolved task, it is promising for real-world usages with the increased data amount. This allows the community to go beyond traditional perception regimens relying on sensor fusion schemes and develop novel methods addressing transparent object segmentation. For example, AdaptiveASPP~\cite{fakemix} is designed to extract rich features of multiple fields-of-view with appropriate importances, whereas EBLNet~\cite{enhanced_boundary_learning} incorporates an edge-aware graph convolution module to model global shape representations.
Differing from most of these accuracy-oriented methods, we aim for a both efficient and robust semantic segmentation desirable for navigation assistance. We establish a transformer-based system to assist the detection of transparent objects in real-world scenes.

\noindent\textbf{Efficient transformers for dense prediction.}
Due to the capacity to model long-range contextual correlations, attention in transformers~\cite{vaswani2017attention} has been introduced in visual recognition tasks to learn inter-dependencies either in the channel or in the spatial dimension~\cite{danet,nonlocal,ocnet} by appending attention layers atop convolutional networks. To reduce the quadratic computation overhead of such non-local attention layers \wrt the input size, their disentangled or asymmetric versions~\cite{ccnet,yang2021capturing,yin2020disentangled,zhu2019asymmetric} are constructed.

Recently, transformers are directly applied in vision tasks~\cite{detr,vit,deformable_detr}. In ViT~\cite{vit} and DeiT~\cite{deit}, a pure transformer is utilized to sequences of image patches for image recognition.
For pixel-wise tasks, SETR~\cite{setr} views semantic segmentation from a sequence-to-sequence perspective with vision transformers~\cite{vit}, whereas MaX-DeepLab~\cite{maxdeeplab} infers class masks with a dual-path transformer for panoptic segmentation. Inspired by their success, transformer architectures for dense prediction emerge~\cite{twins,swin,segmenter,fully_transformer_networks,p2t,segformer}.

A vital set of these models is proposed with lightweight variants like Pyramid Vision Transformers (PVT)~\cite{pvt}, ResT~\cite{rest}, and LeViT~\cite{levit}, aiming to optimize the accuracy-efficiency trade-off when porting transformers to real-world applications. In this work, we devise an efficient \emph{Trans4Trans} framework with focus set on assisting navigation of visually impaired people in the wild. In contrast to existing works that either stack attention layers~\cite{danet,yang2021capturing} and encoder-decoder transformers on CNN backbones~\cite{xie2021segmenting}, or employ CNN-based decoders on top of transformer encoders~\cite{pvt,setr}, in \emph{Trans4Trans} both encoder and decoder are based on transformers, together with a novel Transformer Parsing Module design in our dual-head decoder.



\begin{figure*}[!t]
    \centering
    \includegraphics[width=0.99\linewidth]{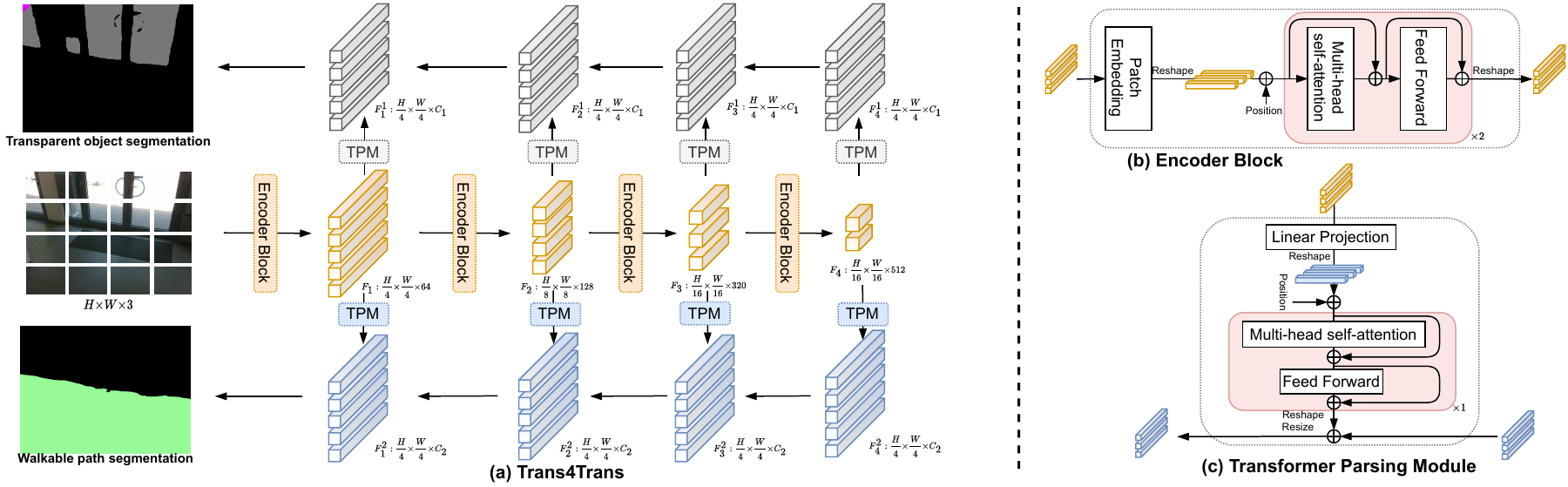}
    \vskip-2ex
    \caption{\small The architecture of (a) Trans4Trans model consists of shared encoder and dual decoders, while (b) and (c) are the general transformer-based encoder block and our proposed Transformer Parsing Module~(TPM) for decoder, respectively.}
    \label{fig:arch_trans4trans}
    \vskip-3ex
\end{figure*}

\section{System Architecture}

\subsection{Trans4Trans}
Inspired by the benefit of the ViT~\cite{vit} transformer model in acquiring long-range dependencies, our dual-head Trans4Trans model is entirely composed of transformers, as shown in Fig.~\ref{fig:arch_trans4trans}(a), while the single-head has only one decoder. The four-stage encoder is borrowed from PVT~\cite{pvt}.
Different to PVT-based Trans2Seg~\cite{xie2021segmenting} adopting CNN-decoder, both encoder and decoder of Trans4Trans are symmetrically constructed by transformers for maintaining consistency in both feature extraction and feature parsing stages. Furthermore, different from CNN-based models~\cite{espnetv2,fastscnn,bisenet,ocnet} learning the inductive bias, the transformer-based decoder is supposed to be more robust to parse unseen data captured in the wild. Yet, training a transformer model requires a large-scale dataset~\cite{vit}. In order to solve the data-hunger problem and correct the misidentified walkable area through transparent objects segmentation, we designed a double-head model. Through the joint training on multiple datasets, it brings greater data diversity for learning a robust transformer-based model.

To construct such a lightweight decoder, we propose a \emph{Transformer Parsing Module~(TPM)} illustrated on Fig.~\ref{fig:arch_trans4trans}(c). Each TPM contains only one single transformer-based layer, thus it only demands little computing resources and is flexible to be deployed on our portable hardware system. More precisely, our decoder consists of four symmetrical stages as encoder. Each stage has a TPM module and contains similar structure. As shown in Fig.~\ref{fig:arch_trans4trans}(a), the pyramid features $\{F_1, F_2, F_3, F_4\}$ from encoder are parsed consistently by the specific TPM module. Between two stages, resize and element-wise addition will be performed for pyramid feature fusion. For balancing the capacity and computational demands, the feature resolution of each TPM is set as $\frac{H}{4}\times{\frac{W}{4}}\times{C}$, for which the default channel number is $64$.

Benefiting from our proposed TPM, the amount of GFLOPs and parameters of this dual-head structure is largely reduced compared to deploying two separate models. Also importantly, diverse features can be learned from various datasets. Thereby, the dual-head model maintains lightweight and is robust in terms of preventing overfitting when testing in real-world scenarios. The decoder composed of our TPM module can be flexibly applied with various CNN- or transformer-based encoder structures as well. For multi-task learning, mounting decoder heads robustifies the feature learned via the shared encoder, and the entire model will not be computationally overburdened.

\subsection{Portable System}
Our entire portable system consists of two hardware components: a pair of smart vision glasses and a portable GPU, \eg, NVIDIA AGX Xavier or a lightweight laptop.

The smart vision glasses integrate a RealSense R200 RGB-D sensor to capture RGB and depth images at the resolution of $640{\times}480$ in real time, and a pair of bone-conducting earphones for generating acoustic feedback to visually impaired people.
This is critical as visually impaired people often use the sounds from their surrounding environments for orientation and bone-conducting headphones will not block their ears when using the system. 
The integrated RealSense R200 sensor leverages a combination of active speckle projecting and passive stereo matching, and thereby it can work in both indoor and outdoor scenes.
In texture-less indoor scenes, the projected infrared speckles will augment the environments, which are beneficial for stereo matching algorithms to yield dense depth estimation.
In sunny outdoor scenes, whereas projected patterns would be overwhelmed by sunlight, the infrared components of natural light shine on the scene to form well-textured infrared image pairs, thus producing robust depth sensing.
In our system, depth information is used to perform the obstacle avoidance function and can be used for prioritizing near-range objects over mid-/long-range objects.

\subsection{System Algorithm}
Our software components are the aforementioned dual-head Trans4Trans and a user interface as described in Algorithm~\ref{algo:system}.
Starting from the input data and to guarantee the timely capture of the facing environment, the frame rate of RGB-D stream is set to $60$. Once the system starts, it repeats image segmentation every $n$ seconds. According to our experiments, the time interval setting as $2$ seconds can effectively prevent cognitive overload, especially in cases of complex scenes containing a large number of objects. Still, it is adjustable depending on the demands of users, \eg, a short interval for more feedback to explore unknown space.

\begin{algorithm}[!t]
	\caption{Assistive system}
	\label{algo:system}
	\KwData{RGB-D as ${X}\in \mathcal{R}^{H \times W \times 3}$ and ${Y}\in \mathcal{R}^{H \times W}$.}
	\KwResult{General segmentation $G\in \mathcal{R}^{{H}\times{W}\times{13}}$; Transparency $T \in \mathcal{R}^{{H}\times{W}\times{11}}$; }
    initialize walkable rate: $R_{l}, R_{f}, R_{r}$, parameters: $\theta_{obstacle}$, $\theta_{trans}$, $\theta_{walkable}$ \;
    \While{system start and each $n$ seconds }{
    RGB-D update and Trans4Trans segmentation: \\
    $G_{path}\in \mathcal{R}^{H \times W}, G_{object}\in \mathcal{R}^{H \times W \times 12}$ \;
    ${T_{stuff}\in \mathcal{R}^{H \times W \times 3}, T_{thing}\in \mathcal{R}^{H \times W \times 8}}$ \;
    partition $\{R_{l}, R_{f}, R_{r}\} \leftarrow G_{path}$ \;
    \uIf{$\overline{Y} < \theta_{obstacle}$}{
    $vibration$ as obstacle warning\;
    }
    \uElseIf{$max\{\overline{T}_i\}\in T_{stuff} > \theta_{trans}$}{
    $speech \leftarrow argmax\{\overline{T}_i\}\in T_{stuff}$ ;
    }
    \uElseIf{$max\{R_{l}, R_{f}, R_{r}\} > \theta_{walkable}$}{
    $speech \leftarrow argmax\{R_{l}, R_{f}, R_{r}\} \in \{left, forward, right\}$;
    }
    \Else{
    $speech \leftarrow nearest\{T_{thing}, G_{object}\}$\;
    }
    }
\end{algorithm}
\noindent\textbf{Obstacle avoidance.}
When moving in a relatively restricted indoor space, the building materials or densely-arranged objects will impede the flexibility of merely using white cane as the aid tool for avoiding obstacles. In order to tackle the collision issue and balance indoor and outdoor scenarios, our system presets the highest priority for obstacle avoidance. In other words, if the average value of the depth information is smaller than the preset distance threshold $\theta_{obstacle}$, the user will be immediately notified in the form of \emph{vibration}. To minimize the uncertainty of vibrations and the cognitive load, only one single default threshold is set to $1$ meter, instead of setting various vibration frequencies for different distances. Another purpose is to preclude the chaotic and low-confidence segmentation from the less-textured images when users walk too close and face to the object surface, such as images from white wall or doors.

\noindent\textbf{(Transparent) object segmentation.}
After receiving the RGB image ${X}\in \mathcal{R}^{H{\times}W{\times}3}$, our efficient Trans4Trans model outputs two segmentation predictions, which are general object segmentation $G\in \mathcal{R}^{H{\times}W{\times}{13}}$ and transparent object segmentation $T \in \mathcal{R}^{H{\times}W{\times}{11}}$, respectively. The general object segmentation is divided into $G_{path}$ for \emph{walkable path} and $G_{object}$ for other \emph{objects}. Afterwards, the walkable mask is further partitioned into three regions as $\{left, forward, right\}$ directions for orientation. In order to correct the wrongly-segmented walkable area by the high-confidence transparency perception, the transparent object segmentation is divided into two disjoint sets as: $T_{stuff}\in \mathcal{R}^{H{\times}W{\times}3}$ with $\{$\emph{window, glass door, glass wall}$\}$, and $T_{things}\in \mathcal{R}^{H{\times}W{\times}8}$ with $\{$\emph{shelf, jar/tank, freezer, eyeglass, cup, bowl, bottle, box}$\}$.

\noindent\textbf{Walkable path detection.}
After achieving object segmentation, the local ratio of walkable area $G_{path}$, \eg, \emph{floor} category from Stanford2D3D, is further horizontally divided into three different directions as $\{R_{l}, R_{f}, R_{r}\} \leftarrow G_{path}$. Then, an intuitive and effective strategy is to prompt the direction that has the largest walkable area, only when its local ratio is greater than the preset threshold $\theta_{walkable}$ for safety. According to our test, this orientation approach guarantees anti-veering in a straight path outdoors and indoors. Furthermore, it can also accurately predict the best instantaneous turning direction during walking at an intersection, so as to constantly yield a safe direction suggestion.

\section{Experiments}

\subsection{Datasets and settings}
\noindent\textbf{Trans10K-v2}~\cite{xie2021segmenting} contains 10,428 images with a $835{\times}1113$ resolution and 11 object categories: \emph{shelf, jar or tank, freezer, window, glass door, eyeglass, cup, wall, glass bow, water bottle, storage box}. It is divided as 5,000, 1,000 and 4,428 images for training, validation and testing.

\noindent\textbf{Stanford2D3D}~\cite{stanford2d3d} contains 70,496 images with a $1080{\times}1080$ resolution and 13 object categories: \emph{beam, board, bookcase, ceiling, chair, clutter, column, door, floor, sofa, table, wall, window}. We follow the fold-1 setting as~\cite{stanford2d3d} which splits Area 1, 2, 3, 4 and 6 as training set, Area 5a and 5b as validation and testing set.

\noindent\textbf{Implementation details.} We implement the model with PyTorch 1.8.0 and CUDA 11.2. Learning rate is initialized as $1e-4$ and is scheduled by poly strategy~\cite{bisenet} with power 0.9 in 100 epochs. The Adam~\cite{adam_optimization} with epsilon $1e-8$ and weight decay $1e-4$ is used as the optimizer. Batch size is set as 4 on each of four 1080Ti GPUs. To maintain the shape of position embedding, the images are resized in the resolution of $512{\times}512$ for all experiments. For a fair comparison with~\cite{xie2021segmenting}, some tricks such as OHEM, auxiliary or class-weighted loss are not applied in our experiments.

\subsection{Segmentation accuracy}
In this subsection, we present analysis for the experiments on different datasets and the results for the combination of CNN/Transformer to materialize varied encoder-decoder structures. Experimental results on computation complexity in GFLOPs and segmentation accuracy are presented and compared with state-of-the-art methods.

\begin{table}[!t]
\centering
\resizebox{\columnwidth}{!}{
\begin{tabular}{@{}lllccll@{}}
\toprule
\textbf{Network} & \textbf{Encoder} & \textbf{Decoder} & \textbf{GFLOPs} & \textbf{MParams} & \textbf{Stanford2D3D} & \textbf{Trans10K-v2} \\ \midrule \midrule
Trans2Seg-T     & R18~\cite{resnet}      & Transformer~\cite{xie2021segmenting} &  16.96 & 17.87 & 42.07 & 64.20 \\
Trans2Seg-T     & R34~\cite{resnet}      & Transformer~\cite{xie2021segmenting} &  30.26 & 27.98 & 42.91 & 68.84 \\
Trans2Seg-S     & R50~\cite{resnet} & Transformer~\cite{xie2021segmenting} &  40.98 & 30.53 & 43.83 & 69.20 \\
\midrule 
PVT-T   & PVT-T~\cite{pvt} & Transformer~\cite{xie2021segmenting} &10.16  &13.11 & 41.00 & 64.60    \\
PVT-S   & PVT-S~\cite{pvt} & Transformer~\cite{xie2021segmenting} &19.58  &24.36 & 41.89 & 68.47     \\
PVT-M   & PVT-M~\cite{pvt} & Transformer~\cite{xie2021segmenting} &49.00  &56.20 & 42.49 & 72.10     \\ 
\midrule
Trans4Trans-T & PVT-T~\cite{pvt}    & TPM / Single-head &10.45  &12.71 & 41.28\gbf{+0.28} & 68.63\gbf{+4.03}    \\
Trans4Trans-S & PVT-S~\cite{pvt}    & TPM / Single-head &19.92  &23.95 & 44.47\gbf{+3.04} & 74.15\gbf{+5.68}     \\
Trans4Trans-M & PVT-M~\cite{pvt}    & TPM / Single-head &34.38  &43.65 & 45.73\gbf{+3.24} & 75.14\gbf{+3.04}     \\
\midrule
Trans4Trans-T & PVT-T~\cite{pvt}    & TPM / Dual-head &  11.22 & 13.10 & 40.44\rbf{-0.56} & 69.84\gbf{+5.24}   \\
Trans4Trans-S & PVT-S~\cite{pvt}    & TPM / Dual-head &  20.69 & 24.34 & 43.45\gbf{+1.56} & 74.57\gbf{+6.10}     \\
Trans4Trans-M & PVT-M~\cite{pvt}    & TPM / Dual-head &  35.17 & 44.04 & 45.15\gbf{+2.66} & 74.98\gbf{+2.88}   \\
\bottomrule[1px]
\end{tabular}
}
\vskip-2ex
\caption{\small Comparison with state-of-the-art methods on Stanford2D3D~\cite{stanford2d3d} and Trans10K-v2~\cite{xie2021segmenting}. 
\#MParams, \#GFLOPs are calculated with the input size of $512{\times}512$.
}
\label{tab:performance}
\vskip-3ex
\end{table}

\noindent\textbf{Results.}
As shown in Table~\ref{tab:performance}, four main encoder-decoder structures are utilized for comparison.
Unlike ResNet encoder-based Trans2Seg~\cite{xie2021segmenting} and PVT~\cite{pvt}, our Trans4Trans uses both Transformer-based encoder and decoder with a TPM design.
It can be seen that single-head Trans4Trans-Medium has achieved the best performance in mIoU on both Stanford2D3D ($45.73\%$) and Trans10K-v2 ($75.14\%$), exceeding by more than $3\%$ on the challenging transparent object segmentation benchmark \wrt PVT-Medium. Meanwhile, it has clearly the smaller computation complexity in GFLOPs compared to PVT-M and ResNet50-based Trans2Seg.
Trans4Trans-Tiny and -Small also achieve higher performances on Trans10K-v2 than the state-of-the-art structures.
Dual-head Trans4Trans consistently improves the performance on Trans10K-v2 by incorporating more general knowledge when learning jointly with supervision from Stanford2D3D, which is more suitable for real-world navigational perception, as it highly prevents overfitting and reduces false positives of transparent obstacle warning observed in our field tests. Overall, these results verify the superiority and efficiency of Trans4Trans for transparent and general object segmentation.

\noindent\textbf{Combination of CNN/Transformer.}
As shown in Table~\ref{tab:combination}, varied combinations of CNN-/Transformer-based encoder and decoder are compared, where FCN~\cite{fcn} and OCNet~\cite{ocnet} are composed of only CNN, whereas Trans2Seg is composed of CNN-based encoder and transformer-based decoder. The proposed Trans4Trans is a fully transformer-based encoder-decoder structure. It outperforms both these competitive architectures and PVT, another transformer-based encoder-decoder architecture. Yet, our Trans4Trans keeps smaller GFLOPs while being more accurate, demonstrating its suitability for transparent object segmentation.

\begin{table}[ht]
\small
\centering
\resizebox{\columnwidth}{!}{
\begin{tabular}{c|cc|cc|cc}
\toprule
\textbf{Method} & \textbf{Trans. Enc.} & \textbf{CNN Enc.} & \textbf{Trans. Dec.} & \textbf{CNN Dec.} & \textbf{GFLOPs}$\downarrow$ & \textbf{mIoU}~(\%)~$\uparrow$ \\ 
\midrule
FCN~\cite{fcn}                      &                &  \checkmark   &               & \checkmark   & 42.2  & 62.7 \\ 
OCNet~\cite{ocnet}                  &                &  \checkmark   &               & \checkmark   & 43.3 & 66.3 \\ 
Trans2Seg~\cite{xie2021segmenting}  &                &  \checkmark   & \checkmark    &              & 40.9 & 69.2 \\ 
PVT~\cite{pvt}                      & \checkmark     &               & \checkmark    &              & 49.0 & 72.1 \\
Trans4Trans (ours)                  & \checkmark     &               & \checkmark    &              & \textbf{34.3} & \textbf{75.1} \\
\bottomrule
\end{tabular}
}
\vskip-1ex
\caption{\small Effect of CNN/Transformer combination.
Models are evaluated on the Trans10K-v2 dataset.}
\label{tab:combination}
\vskip-3ex
\end{table}

\noindent\textbf{Comparison to state-of-the-art models.}
Following~\cite{xie2021segmenting}, we compare with both accuracy- and efficiency-oriented semantic segmentation models as shown in Table~\ref{tab:sota}.
Compared with both CNNs and transformer-based methods like Trans2Seg~\cite{xie2021segmenting}, the superiority of Trans4Trans is further confirmed.
Our Trans4Trans-M model outperforms the state-of-the-art method Trans2Seg by $2.99\%$ in mIoU and $0.87\%$ in ACC, while requiring much less GFLOPs.
For category-wise accuracy, our Trans4Trans model achieves the state-of-the-art IoU on the classes \emph{background}, \emph{jar or tank}, \emph{window}, \emph{door}, \emph{cup}, \emph{wall}, \emph{bottle} and \emph{box}.
These experimental results show the efficacy of transparent object segmentation of the proposed Trans4Trans architecture.

\begin{table*}[!t]
\footnotesize
\centering
\scalebox{0.85}{
\begin{tabular}{p{80pt}<{\raggedright}p{30pt}<{\centering}p{22pt}<{\centering}p{22pt}<{\centering}p{30pt}<{\centering}p{18pt}<{\centering}p{22pt}<{\centering}p{18pt}<{\centering}p{18pt}<{\centering}p{18pt}<{\centering}p{18pt}<{\centering}p{18pt}<{\centering}p{18pt}<{\centering}p{18pt}<{\centering}p{18pt}<{\centering}p{18pt}<{\centering}}
\toprule
\multirow{2}{*}{\textbf{Method}} & \multirow{2}{*}{\textbf{GFLOPs}~$\downarrow$} & \multirow{2}{*}{\textbf{ACC}~$\uparrow$} & \multirow{2}{*}{\textbf{mIoU}~$\uparrow$} & \multicolumn{12}{c}{\rule{0pt}{10pt}\textbf{Category IoU}~$\uparrow$} \\ \cline{5-16} 

\rule{0pt}{10pt} &  \multicolumn{1}{c}{} &  &  \multicolumn{1}{c}{} & Background & Shelf  & Jar/Tank  & Freezer & Window & Door & Eyeglass & Cup & Wall & Bowl & Bottle & Box  \\ \midrule
 
\rule{0pt}{10pt} FPENet~\cite{fpenet}  &\multicolumn{1}{|c|}{0.76} & 70.31 & \multicolumn{1}{c|}{10.14} & 74.97 & 0.01 & 0.00 & 0.02 & 2.11 & 2.83 & 0.00 & 16.84 & 24.81 & 0.00 & 0.04 & 0.00 \\
\rule{0pt}{10pt} ESPNetv2~\cite{espnetv2}  & \multicolumn{1}{|c|}{0.83} & 73.03 & \multicolumn{1}{c|}{12.27} & 78.98 & 0.00 & 0.00 & 0.00 & 0.00 & 6.17 & 0.00 & 30.65 & 37.03 & 0.00 & 0.00 & 0.00 \\
\rule{0pt}{10pt} ContextNet~\cite{contextnet} & \multicolumn{1}{|c|}{0.87} & 86.75 & \multicolumn{1}{c|}{46.69}   & 89.86 & 23.22 & 34.88 & 32.34 & 44.24 & 42.25 & 50.36 &  65.23 & 60.00 & 43.88 & 53.81 & 20.17 \\
\rule{0pt}{10pt} FastSCNN~\cite{fastscnn} & \multicolumn{1}{|c|}{1.01} & 88.05 & \multicolumn{1}{c|}{51.93} & 90.64 & 32.76 & 41.12 & 47.28 &  47.47 & 44.64 & 48.99 & 67.88 & 63.80 & 55.08 & 58.86 & 24.65 \\
\rule{0pt}{10pt} DFANet~\cite{dfanet}  & \multicolumn{1}{|c|}{1.02} & 85.15 & \multicolumn{1}{c|}{42.54} & 88.49 & 26.65 & 27.84 & 28.94 & 46.27 & 39.47 & 33.06 & 58.87 & 59.45 & 43.22 & 44.87 & 13.37 \\
\rule{0pt}{10pt} ENet~\cite{enet}  & \multicolumn{1}{|c|}{2.09} & 71.67 & \multicolumn{1}{c|}{8.50} & 79.74 & 0.00 &  0.00 & 0.00 & 0.00 & 0.00 & 0.00 & 0.00 & 22.25 & 0.00 & 0.00 & 0.00 \\
\rule{0pt}{10pt} DeepLabv3+MBv2~\cite{mobilenetv2} & \multicolumn{1}{|c|}{2.62} & 88.39 & \multicolumn{1}{c|}{54.16} & 89.95 & 31.79  & 48.29  & 46.18 & 41.39 & 43.42 & 61.97 & 69.48 & 61.65 & 54.89 & 63.47 & 37.36  \\
\rule{0pt}{10pt} HRNet\_w18~\cite{hrnet}  & \multicolumn{1}{|c|}{4.20} & 89.58 & \multicolumn{1}{c|}{54.25} & 92.47 & 27.66 & 45.08 & 40.53 & 45.66 & 45.00 & 68.05 & 73.24 & 64.86& 52.85 & 62.52 & 33.02   \\
\rule{0pt}{10pt} HarDNet~\cite{hardnet}  & \multicolumn{1}{|c|}{4.42} & 90.19 & \multicolumn{1}{c|}{56.19} & 92.87 & 34.62 & 47.50 & 42.40 & 49.78 & 49.19 & 62.33 & 72.93 & 68.32 & 58.14 & 65.33 & 30.90 \\
\rule{0pt}{10pt} DABNet~\cite{dabnet} & \multicolumn{1}{|c|}{5.18} & 77.43 & \multicolumn{1}{c|}{15.27} & 81.19 & 0.00 & 0.09 & 0.00 & 4.10 & 10.49 & 0.00 & 36.18 & 42.83& 0.00 & 8.30 & 0.00  \\
\rule{0pt}{10pt} LEDNet~\cite{lednet}  & \multicolumn{1}{|c|}{6.23} & 86.07 & \multicolumn{1}{c|}{46.40} & 88.59 & 28.13 & 36.72 & 32.45 & 43.77 & 38.55 & 41.51 & 64.19 & 60.05 & 42.40 & 53.12 & 27.29 \\
\rowcolor{gray!15}\rule{0pt}{10pt} Trans4Trans-T &  \multicolumn{1}{|c|}{10.45} &  93.23  & \multicolumn{1}{c|}{68.63} & 94.44 & 48.39 & 61.89 & 61.86 & 61.14 & 54.83 & 73.60 & 83.03 & 75.20 & 74.69 & 75.26 & 59.19  \\ 
\midrule

\rule{0pt}{10pt} ICNet~\cite{icnet}  & \multicolumn{1}{|c|}{10.64} & 78.23 &\multicolumn{1}{c|}{23.39} & 83.29 & 2.96 & 4.91 & 9.33 & 19.24 & 15.35 & 24.11 & 44.54 & 41.49 & 7.58 & 27.47 & 3.80  \\
\rule{0pt}{10pt} BiSeNet~\cite{bisenet}  & \multicolumn{1}{|c|}{19.91} & 89.13 & \multicolumn{1}{c|}{58.40} & 90.12 & 39.54 &53.71  &50.90  & 46.95& 44.68 & 64.32 & 72.86 &63.57 &61.38 & 67.88 &44.85  \\
\rowcolor{gray!15}\rule{0pt}{10pt} Trans4Trans-S &  \multicolumn{1}{|c|}{19.92} &  94.57  & \multicolumn{1}{c|}{74.15} & 95.60 & \textbf{57.05} & 71.18 & \textbf{70.21} & 63.95 & 61.25 & 81.67 & 87.34 & 78.52 & \textbf{77.13} & 81.00 & 64.88 \\ 
\midrule

\rule{0pt}{10pt} DenseASPP~\cite{denseaspp} & \multicolumn{1}{|c|}{36.20} & 90.86 & \multicolumn{1}{c|}{63.01} & 91.39 & 42.41 & 60.93 & 64.75 & 48.97 & 51.40 & 65.72 & 75.64 & 67.93 & 67.03 & 70.26 & 49.64  \\
\rule{0pt}{10pt} DeepLabv3+~\cite{deeplabv3+} & \multicolumn{1}{|c|}{37.98} & 92.75 & \multicolumn{1}{c|}{68.87}  & 93.82 & 51.29 &64.65  & 65.71 & 55.26 & 57.19 & 77.06 & 81.89 & 72.64 &70.81  & 77.44 & {58.63}\\
\rule{0pt}{10pt} FCN~\cite{fcn}  & \multicolumn{1}{|c|}{42.23} & 91.65 & \multicolumn{1}{c|}{62.75} & 93.62 & 38.84 & 56.05 & 58.76 & 46.91 & 50.74 & 82.56 &78.71 & 68.78 & 57.87 & 73.66 & 46.54 \\
\rule{0pt}{10pt} OCNet~\cite{ocnet}  & \multicolumn{1}{|c|}{43.31} & 92.03 & \multicolumn{1}{c|}{66.31} & 93.12 & 41.47 & 63.54 & 60.05 & 54.10 & 51.01 & 79.57 & 81.95 & 69.40 & 68.44 & 78.41 & 54.65\\
\rule{0pt}{10pt} RefineNet~\cite{refinenet}  & \multicolumn{1}{|c|}{44.56} & 87.99 & \multicolumn{1}{c|}{58.18} & 90.63 & 30.62 & 53.17 & 55.95 & 42.72 &46.59 & 70.85 & 76.01 & 62.91 & 57.05 & 70.34 & 41.32   \\
\rule{0pt}{10pt} Trans2Seg~\cite{xie2021segmenting} &  \multicolumn{1}{|c|}{49.03} &  {94.14}  & \multicolumn{1}{c|}{{72.15}} & {95.35} & 53.43 & {67.82} & 64.20 & {59.64} & {60.56} & \textbf{88.52} & {86.67} & {75.99} & {73.98} & {82.43} & 57.17 \\ 
\rule{0pt}{10pt} TransLab~\cite{xie2020segmenting}  & \multicolumn{1}{|c|}{61.31} & 92.67  & \multicolumn{1}{c|}{69.00} & 93.90 & {54.36} & 64.48 & 65.14 & 54.58 & 57.72 & 79.85 & 81.61 &72.82 & 69.63 & 77.50 & 56.43 \\ 
\rule{0pt}{10pt} DUNet~\cite{dunet}  & \multicolumn{1}{|c|}{123.69} & 90.67 & \multicolumn{1}{c|}{59.01} & 93.07 & 34.20 & 50.95 & 54.96 & 43.19 & 45.05 & 79.80 & 76.07 & 65.29 & 54.33 & 68.57 & 42.64   \\
\rule{0pt}{10pt} U-Net~\cite{unet}  & \multicolumn{1}{|c|}{124.55} & 81.90 & \multicolumn{1}{c|}{29.23} & 86.34 & 8.76  & 15.18  & 19.02 & 27.13 & 24.73 & 17.26 & 53.40 & 47.36 & 11.97 & 37.79 & 1.77 \\
\rule{0pt}{10pt} DANet~\cite{danet}  & \multicolumn{1}{|c|}{198.00} & 92.70  &\multicolumn{1}{c|}{68.81} & 93.69 & 47.69 & 66.05  & 70.18 & 53.01 & 56.15 & 77.73 & 82.89 & 72.24 & 72.18 & 77.87 & 56.06 \\
\rule{0pt}{10pt} PSPNet~\cite{pspnet}  & \multicolumn{1}{|c|}{187.03} & 92.47 & \multicolumn{1}{c|}{68.23} & 93.62  & 50.33 & 64.24 & {70.19} & 51.51 & 55.27 & 79.27 & 81.93 & 71.95 &   68.91 & 77.13 & 54.43 \\

\midrule 
\rowcolor{gray!15}\rule{0pt}{10pt} Trans4Trans-M &  \multicolumn{1}{|c|}{34.38} &  \textbf{95.01}  & \multicolumn{1}{c|}{\textbf{75.14}} & \textbf{96.08} & 55.81 & \textbf{71.46} & 69.25 & \textbf{65.16} & \textbf{63.96} & 83.84 & \textbf{88.21} & \textbf{80.29} & 76.33 & \textbf{83.09} & \textbf{68.09} \\ 

\bottomrule
\end{tabular}
}
\vskip-1ex
\caption{Computation complexity in GFLOPs and category-wise accuracy evaluation and comparison with state-of-the-art semantic segmentation methods on the Trans10K-v2 dataset~\cite{xie2021segmenting}.}
\label{tab:sota}
\vskip-3ex
\end{table*}

\begin{figure}[t]
    \centering
    \includegraphics[width=0.99\linewidth]{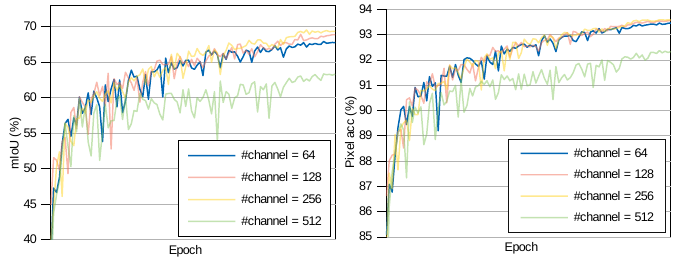}
    \vskip-2ex
    \caption{\small mIoU and pixel accuracy curves under different embedding channels in TPM.}
    \label{fig:channel_test_miou}
    \vskip-4ex
\end{figure}

\noindent\textbf{Channel of TPM.}
Since one of our critical designs lies in the TPM, we now analyze the effect of the numbers of embedding channels applied in the decoder of Trans4Trans, as shown in Table~\ref{tab:channel_decoder}.
It can be seen that performance increases as the number of channels increases until $256$, and it drops at $512$ where the decoder overfits the encoded feature (see Fig.~\ref{fig:channel_test_miou}) and the computation complexity becomes exceedingly large.
For the response time-critical wearable system, we adopt $64$ channels when deploying Trans4Trans due to its high efficiency and good performance.

\begin{table}[t]
\centering
\resizebox{.85\columnwidth}{!}{
\begin{tabular}{c|cccc}
    \toprule
    {\textbf{Channel}}&{\textbf{MParams}}&{\textbf{GFLOPs}}&\textbf{Acc~(\%)}&\textbf{mIoU~(\%)}\\
    \midrule
    64     & 10.45 & 12.71 & 93.46 & 67.89 \\
    128    & 12.46 & 14.02 & 93.49 & 68.88 \\
    256    & 20.41 & 17.82 & 93.58 & 69.51 \\
    512    & 51.50 & 33.82 & 92.37 & 63.33 \\
    \bottomrule
\end{tabular}}
\vskip-2ex
\caption{\small Effect of embedding channel in TPM. All tiny Trans4Trans are trained on Trans10K-v2 at $512{\times}512$ on one GPU.}
\label{tab:channel_decoder}
\vskip-3ex
\end{table}

\subsection{Real-time performance}
To calculate the inference speed of our different versions of dual-head Trans4Trans model, 300 samples from the Trans10K-v2 test set with a batch size of 1 and a resolution of $512{\times}512$ are tested on three different GPUs, \ie, a mobile NVIDIA AGX Xavier in the MAXN mode, an NVIDIA GeForce MX350 from a lightweight laptop and an RTX 2070 from a workstation. As shown in Table~\ref{tab:inf_time}, the computation costs of our tiny Trans4Trans model on three GPUs are considerably lower than the other two, meanwhile the performances of the three models on both datasets are suitable for our system. In real applications, the more timely response of the navigation system is beneficial for assisting users with a similar prediction accuracy on each frame.
Hence, the tiny version is selected in our user study. 

\begin{table}[h]
\centering
\resizebox{\columnwidth}{!}{
\begin{tabular}{l|ccc}
    \toprule
    {\textbf{Network}}&\textbf{NVIDIA Xavier~(ms)~$\downarrow$}&{\textbf{MX350~(ms)~$\downarrow$}}&{\textbf{RTX 2070~(ms)~$\downarrow$}}\\
    \midrule
    Trans4Trans-M & 115.9(\textpm 1.1) / 202.8(\textpm 1.1)& 186.1(\textpm 0.3) / 243.2(\textpm 0.3) & 22.9(\textpm 0.3) / 36.6(\textpm 0.8) \\
    Trans4Trans-S & 95.3(\textpm 0.6) / 158.6(\textpm 1.8)& 140.6(\textpm 0.3) / 188.4(\textpm 0.4) & 17.1(\textpm 0.3) / 27.7(\textpm 0.5) \\
    Trans4Trans-T & 75.8(\textpm 0.7) / 122.7(\textpm 0.7) & 101.5(\textpm 0.3) / 141.7(\textpm 1.6) & 12.8(\textpm 0.5) / 20.3(\textpm 0.5)  \\
    
    \bottomrule
\end{tabular}}
\vskip-2ex
\caption{\small Inference time~(ms/frame) of dual-head Trans4Trans is tested in half-/single-precision on various GPUs at $512{\times}512$.}
\label{tab:inf_time}
\vskip-3ex
\end{table}

\subsection{Qualitative segmentation analysis}

\begin{figure}[t]
    \centering
    \includegraphics[width=0.99\linewidth]{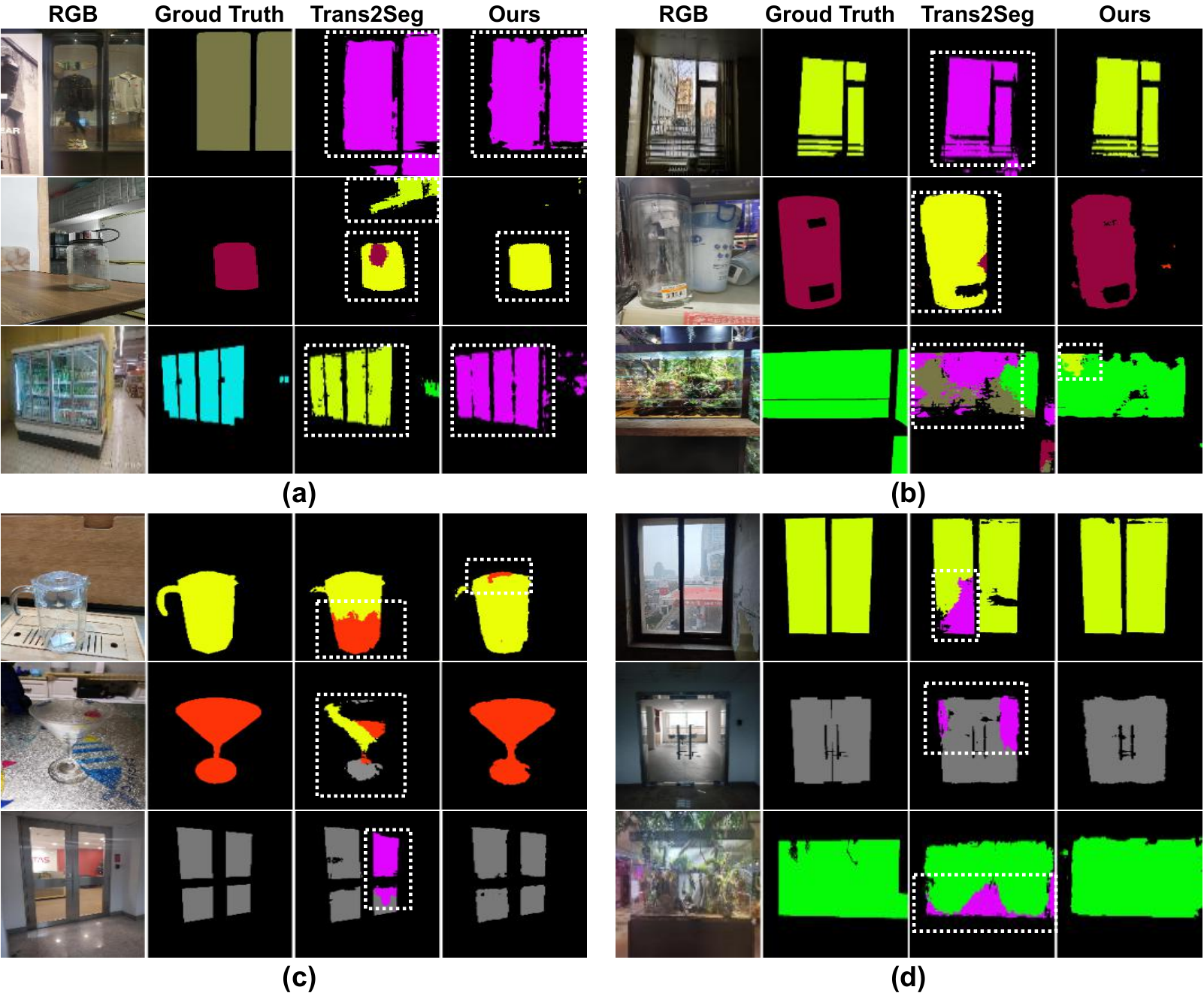}
    \vskip-1ex
    \caption{\small Qualitative analysis on Trans10K-v2 test set. (a) shows some negative predictions from both models. In (b), our Trans4Trans can correctly segment those cases failed by Trans2Seg. In (c) and (d), our results are more precise.
    }
    \label{fig:vis_testset}
    \vskip-3ex
\end{figure}

Fig.~\ref{fig:vis_testset} visualizes qualitative comparisons between our tiny Trans4Trans and the previous state-of-the-art method Trans2Seg~\cite{xie2021segmenting}. Fig.~\ref{fig:vis_testset}(a) shows failed recognition cases of both models, but our model can yield a clearly better boundary. Fig.~\ref{fig:vis_testset}(b) shows that examples where our model predicts the correct label, whereas Trans2Seg is confused. In Fig.~\ref{fig:vis_testset}(c)(d), it can be seen that our model is not only effective for detecting navigation-related \emph{glass door} and \emph{glass window}, but can also predict more refined segmentation of small objects like \emph{jar/tank} and \emph{glass cup}.

\begin{figure}[!t]
    \centering
    \includegraphics[width=0.99\linewidth]{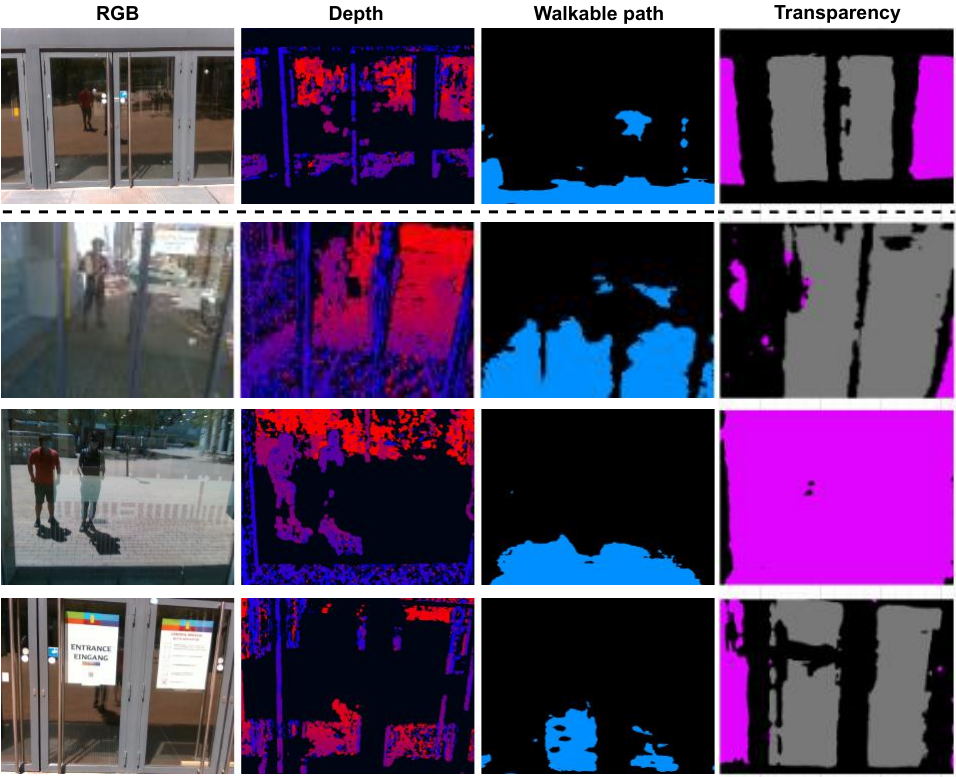}
    \vskip-1ex
    \caption{\small Visualization of real-world scenes. From left to right are RGB and depth image, segmentation as \textcolor{blue}{walkable path} by single-head model trained on Stanford2D3D,
    and as transparent objects (\textcolor{gray}{glass door} or \textcolor{red}{glass wall}) corrected by our dual-head Trans4Trans model.}
    \label{fig:vis_realworld}
    \vskip-3ex
\end{figure}

We further perform field tests by navigating around the university campus and capturing real-world scenes with our smart vision glasses. The collected RGB-D images and corresponding predictions are shown in Fig.~\ref{fig:vis_realworld}.
The glass door in the first row captured at a moderate distance can be correctly identified, whereas in the other rows they are mis-classified as walkable paths by general object segmentation models.
As it can be observed, transparent surfaces are often texture-less and the infrared patterns projected by the glasses will transmit the glass regions, and thereby the depth information are often sparse, noisy or even lost, which makes it challenging for 3D vision-based systems~\cite{aladren2014navigation,wang2017enabling,yang2017ir} to help avoid hazards even though the obstacles are close.
In contrast, our Trans4Trans accurately and completely segments those transparent objects, meanwhile covers general objects, which is ideally suitable for safety-critical navigation assistance.

\section{User Study}
We conducted a qualitative study with 5 participants in order to assess the acceptance of our prototype and draw design conclusions \cite{nielsen1994estimating,nielsen2012howmany}.

\begin{figure}[b]
    \centering
    \includegraphics[width=0.99\linewidth]{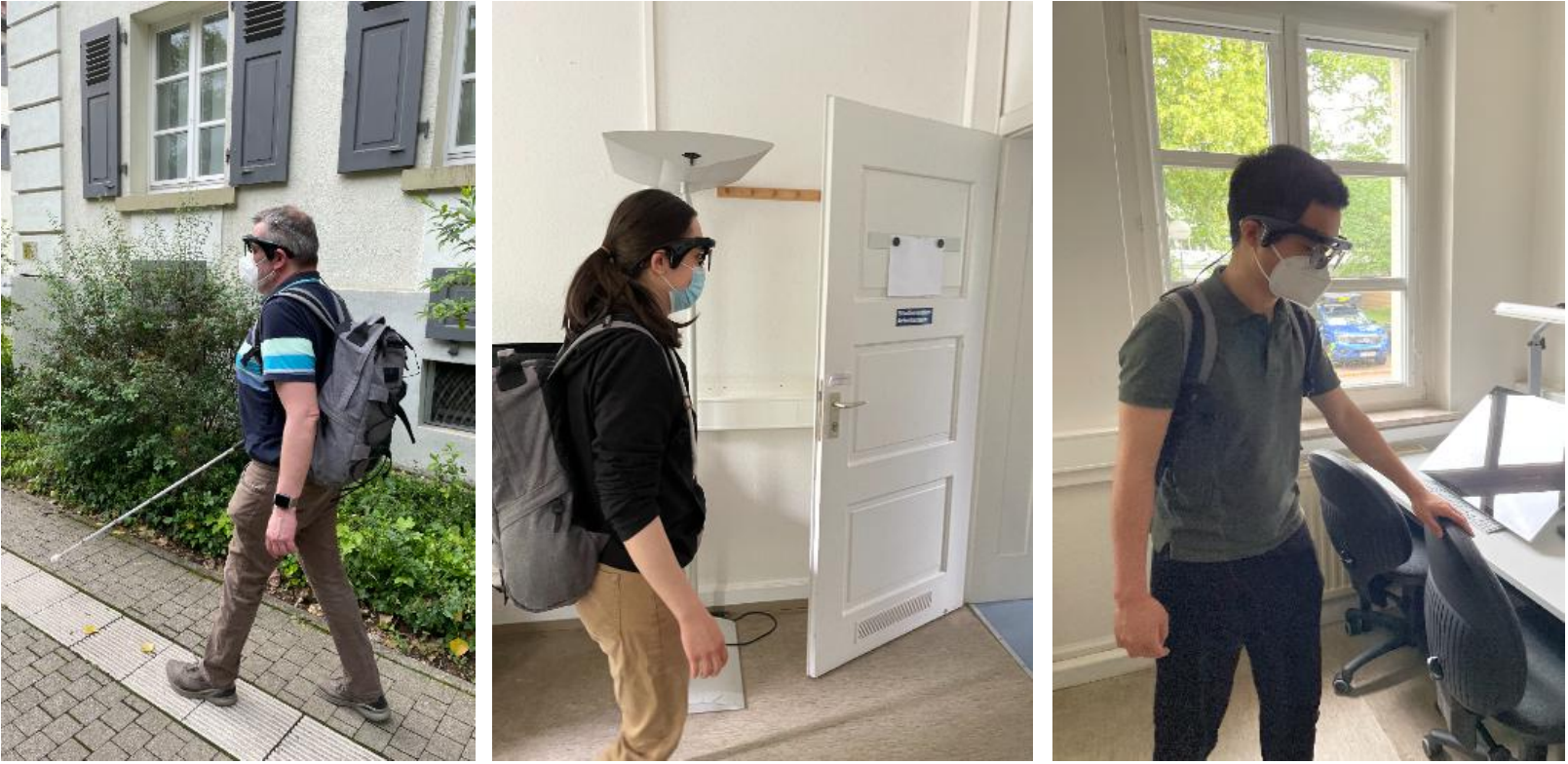}
    \vskip-1ex
    \caption{\small Incidences of participants using the system for navigation outdoors and indoors.}
    \label{fig:participants}
\end{figure}

\noindent\textbf{Methodology.}
The hardware used during the test consisted of the smart vision glasses and 
a backpack with a light-weight laptop and a battery pack inside. The system's battery life under these conditions was approximately $4$ hours.
Participants tried the system inside $2$ buildings, and the blind participant also
on a $700 m$ route outdoors - see Fig.~\ref{fig:participants}.
The study lasted about $2$ hours.
As Corona-protective measures, everyone wore FFP2 or surgery masks throughout the
study
and the prototype was disinfected several times.

After a short introduction, all participants agreed to participation and recordings and signed the data protection statement. The participants put on the system and walked around the rooms, thinking out loud \cite{johnstone2006using}.
The study was recorded with an action camera and voice recorder. At the end, demographics and NASA Raw Task Load Index (RTLX)~\cite{hart2006nasa} questionnaires were filled in.

\noindent\textbf{Participants.} Due to COVID-19 restrictions, only one of the participants (P1B) belonged to the target user group, being early blind. The other 4 participants were sighted (P2-P5). Age and gender of the participants were fairly balanced - see Table~\ref{tab:demographics}. When asked if they can see glass objects during the day, P1B said he can sometimes see closed windows, due to the
light-dark contrast.
Windows that open inside the room, however, are very dangerous, as one can get serious head injuries (P1B). All sighted participants said they can see glass objects most of the time, but some objects, like bottles and glass cups (P2), glass doors (P3,P5), glass walls and windows (P5), can be challenging under certain light conditions, \eg, backlighting.

\begin{table}[h]
\centering
\resizebox{\columnwidth}{!}{
\begin{tabular}{ c c | c c c c | c} 
\multicolumn{2}{c|}{Gender} & \multicolumn{4}{c|}{Age Range} & Hearing loss \\ 
\hline
Male & Female & 18-25 & 26-35 & 36-45 & 46-55 & No\\ 
\hline
3 & 2 & 1 & 2 & 1 & 1 & 5\\
\end{tabular}
}
\vskip-1ex
\caption{Aggregated demographics of participants.}
\label{tab:demographics}
\vskip-3ex
\end{table}

\begin{figure}[h]
    \centering
    \includegraphics[width=0.70\linewidth]{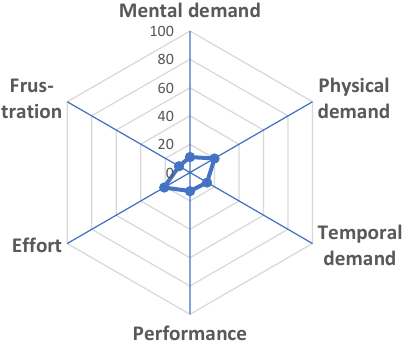}
    \vskip-1ex
    \caption{Cognitive load
    (raw task load index, 0-100).}
    \label{fig:nasa-rtlx}
    \vskip-3ex
\end{figure}

\noindent\textbf{Cognitive load.}
The raw task load index, averaged over all participants, was $14.7$ with a standard deviation of $4.1$. This score is enough to keep the user motivated, while not burdening too much \cite{martinez2020helping}. The blind participant, P1B, had the second lowest score.
According to the individual ratings (Fig.~\ref{fig:nasa-rtlx}), effort and physical demand were slightly higher, while frustration was the lowest subscale. This might suggest that users enjoyed the experience of using our system, but a further reduction of hardware would be welcome.

\noindent\textbf{User comments.}
A thematic analysis~\cite{Braun2006using} performed on the comments made by users yielded the following insights:
\begin{compactitem}
    \item All users found the system useful and were impressed by its functionality (P1B-P5) and smooth running (P5). Users also praised the fact that it works both indoor and outdoor (P2,P3), is easy to use and interact (P2), the several functions are well integrated and battery life is high (P5). P1B said: \emph{``for the first time, I had the feeling that artificial intelligence can be useful [...] It's just cool!''}.
    \item All 5 users liked the fact that the system recognizes so many object classes, including glass objects.
    Objects recommended to be included in future versions were: trash cans, city scouters (P1B) and constructions site fences (P5). Detection of some false positives was mentioned by P1B and P5.  
    \item Users found the hardware light (P1B,P2,P3,P5), comfortable (P1B,P3) and good looking (P5). For a commercial system, however, the laptop should be replaced by a belt (P1B) or smartphone (P5) and the glasses should connect wirelessly (P1B,P5).
    \item The direction of detected objects should be announced (P1B,P2,P5) as clock directions or stereo sound (P1B).
    \item Synthetic voice
    was very much appreciated by P1B, as \emph{``it clearly stands out from background sounds''}.
    \item The $1-2 s$ delay setting was mentioned by P1B,P4,P5. P1B said the delay is acceptable for object segmentation, as he only uses this function when he is unsure.
    P5 only mentioned the delay with respect to obstacle avoidance. P4 said the delay is fine.
\end{compactitem}

\begin{figure}[h]
    \centering
    \includegraphics[width=0.99\linewidth]{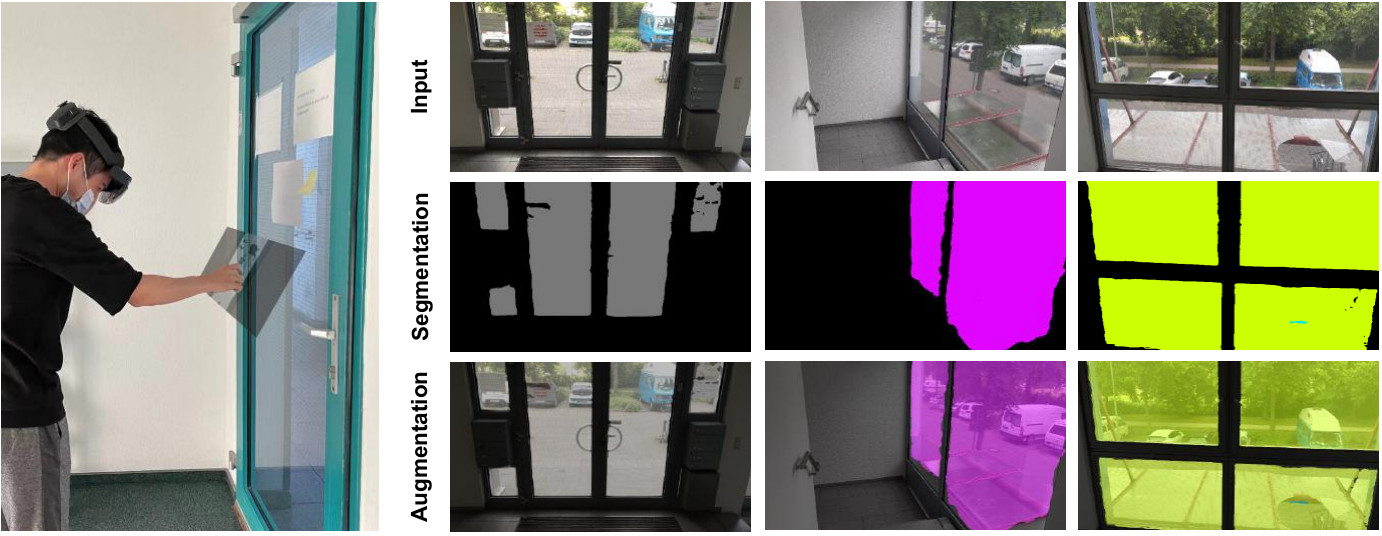}
    \vskip-1ex
    \caption{Augmented results with a HoloLens 2 device.}
    \label{fig:hololens}
    \vskip-3ex
\end{figure}

\noindent\textbf{Augmented reality for partially sighted people.}
Since transparent obstacles are often a threat for people with low vision in everyday navigation and even challenging for sighted people in some cases, we further test our Trans4Trans method with a HoloLens 2 device by capturing real-world data around our computer vision laboratory. As shown in Fig.~\ref{fig:hololens}, the transparent objects like \emph{glass door}, \emph{transparent wall}, and \emph{glass window} can be completely and consistently segmented, and the colored segmentation mask can be easily overlaid and naturally projected onto the original RGB image captured by the glasses for rendering augmented reality or mixed reality. This field test demonstrates that the proposed Trans4Trans framework is not only helpful for assisting blind people, but can be potentially useful for partially sighted people. 

\section{Conclusion}
We look into the perception of transparent object segmentation via Trans4Trans, an efficient transformer architecture established with both transformer-based encoder and decoder. With a novel Transformer Parsing Module (TPM) integrated in the dual-head, Trans4Trans precisely segments general and transparent objects. It attains state-of-the-art performances on Stanford2D3D and Trans10K-v2 datasets, meanwhile being swift and robust to support online navigational perception. The learned efficient vision transformer is ported in our wearable system with a pair of smart vision glasses designed to help visually impaired people travel and explore real-world scenes, where transparent objects are omnipresent. Extensive results from a user study and various field tests show that the proposed assistive system is reliable and with low cognitive load.

\clearpage

{\small
\bibliographystyle{ieee_fullname}
\bibliography{egbib}
}

\end{document}